\documentclass[11pt,a4paper]{article}
\usepackage[hyperref]{naaclhlt2019}
\usepackage{times}
\usepackage{latexsym}
\usepackage{amssymb}
\usepackage{amsmath}
\usepackage[ruled,vlined]{algorithm2e}
\usepackage{kantlipsum}
\usepackage{abstract}
\usepackage{graphicx}
\usepackage{url}
\usepackage{arydshln}
\aclfinalcopy 
\def\aclpaperid{123} 

\title{Meta-Learning for Natural Language Understanding under Continual Learning Framework}

\author{Jiacheng Wang\\
  New York University \\
  {\tt jw5728@nyu.edu}\\\And
  Yong Fan\\
  New York University \\
  {\tt yf869@nyu.edu} \\\And
  Duo Jiang  \\
   New York University \\
  {\tt dj1057@nyu.edu} \\\And
  Shiqing Li\\
  New York University \\
  {\tt sl7085@nyu.edu} \\\
    }

\date{}

\begin{document}
\maketitle
\begin{abstract}
Neural network has been recognized with its accomplishments on tackling various natural language understanding (NLU) tasks. Methods have been developed to train a robust model to handle multiple tasks to gain a general representation of text. In this paper, we implement the model-agnostic meta-learning (MAML) and Online aware Meta-learning (OML) meta-objective under the continual framework for NLU tasks proposed by Javed and White\shortcite{MLRCL:19}. We validate our methods on selected SuperGLUE \shortcite{superglue:19}  and GLUE benchmark \shortcite{glue:19}.
\end{abstract}

\section{Introduction}
One ultimate goal of language modelling is to construct a model like human, to grasp general, flexible and robust meaning in language. One reflection of obtaining such model is be able to master new tasks or domains on same task quickly. However, NLU models have been building from specific task on given data domain but fail when dealing with out-of-domain data or performing on a new task. To combat this issue, several research areas in transfer learning including domain adaptation, cross lingual learning, multi-task learning and sequential transfer learning have been developed to extend model handling on multiple tasks. However, transfer learning tends to favor high-resources tasks if not trained carefully, and it is also computationally expensive \cite{dou:19}.  Meta learning algorithm tries to solve this problem by training model in a variety of tasks which equip the model the ability to adapt to new tasks with only a few samples.

In our case, we adopt the idea of model-agnostic meta learning (MAML) which is an optimization method of meta learning that directly optimized the model by constructing an useful initial representation that could be efficiently trained to perform well on various tasks \cite{finn:18}. However, in an continual learning where data comes into the model sequentially, there is still a potential problem of catastrophic forgetting where a model trained with new tasks would start to perform worse on previous tasks. The two objectives of designing a continual learning architecture are to accelerate future learning where it exploits existing knowledge of a task quickly together with general knowledge from previous tasks to learn prediction on new samples and to avoid interference in previous tasks by updates from new tasks. \cite{MLRCL:19}. 

In this paper, we utilize algorithm derived from Jave and White \shortcite{MLRCL:19} which applies Meta-Learning under continual learning. Our objective is to apply this framework in NLP field, specifically on NLU tasks. By taking advantage of this model-agnostic approach, Meta-Learning under continual learning should be applicable on any language model that is optimized by gradient-based methods. We compare our results with Duo et al \shortcite{dou:19} which applies meta-learning on Glue tasks, our MAML-Rep shows comparable results. We hope to bring new research direction in NLP fields focusing on such method. The implementation of our code can be found at \url{https://github.com/lexili24/NLUProject}. 

\section{Background}
The section is dedicated to examine the implementation of methods solely or combined, in natural language and other fields, which leads us to develop our framework tackling NLU tasks. Plenty of research have been focused in these two areas and some efforts have succeeded in combining these two goals in other field. 

\subsection{Meta-Learning}

There has been success in implementing MAML in NLU tasks \cite{dou:19}. In their work, they explored the model-agnostic meta-learning algorithm (MAML) and its variants for low-resource NLU tasks and obtained impressive results on the GLUE benchmark\cite{glue:19}. This proves that MAML can be applied to NLU tasks, and achieve comparable results on complex architectures like BERT and MT-DNN. However, this method does not address the potential problem of catastrophic forgetting, as they test Meta-Trained model on one task at a time.

In addition, meta learning is proved to excel in other natural language domains. Mi et al\shortcite{mi:19} has shown promising results of incorporating MAML in natural language generation (NLG). NLG models, like many NLU tasks, are heavily affected by the domain they are trained on and are data-intensive but data resource is low due to high annotation cost. Therefore, authors \shortcite{mi:19} approach to generalize a NLG model with MAML to train on the optimization procedure and derive a meaningful initialization serves to adapt new low-resource NLG scenarios efficiently. In comparison, Meta-Learning approach outperformed multi-task approach with higher BLEU score and lower error percentage. 

\subsection{Continual Learning}

Continual learning is proved to boost model performance in Liu et al \shortcite{liu:19}'s writing on computing sentence similarities. Liu et al \shortcite{liu:19} leveraged continual learning to construct a simple linear sentence encoder to learn representations and compute similarities between sentences, such application can be fed into a chat bot. A general concern is that in practice, the encoder is fed into a series of input from inconsistent corpora, and might degrade performance if fails to generalize common knowledge across domains. Continual learning enables zero-shot learning and allows a sentence encoder to perform well on new text domains while avoiding catastrophic forgetting\cite{liu:19}. Authors evaluate result on semantic textual similarity (STS) datasets with Pearson correlation coefficient (PCC). With a structure utilizing continual learning approach, Liu et al \shortcite{liu:19} showed consistent results cross various corpora. 

Continual learning implemented in NLU tasks on top of transfer learning presented by Yogatama \shortcite{Yogatama:2019} did not show generalization of the model. Yogatama et al followed the continual learning setup to train a new task on best SQuAD-trained BERT and ELMo model, and both architectures show catastrophic forgetting after TriviaQA or MNLI is trained, which degrades model performance on SQuAD dataset. Their work shows an attempt to derive a generative language model and provides a solid ground of continual learning in language modelling. 

An implementation of meta-learning under continual framework is proposed in reinforcement learning (RL) by Alshedivat et al\shortcite{Al-Shedivat:2017}. In their paper, MAML is proved to be a complementary solution adding onto continual adaption in reinforcement learning (RL) fields. Al-Shedivat et al\shortcite{Al-Shedivat:2017} considered nonstationary environments as sequences of stationary tasks for RL agents, which transferred nonstationary environment to learning-to-learning tasks. They developed a gradient-based Meta-Learning algorithm for quick adaption to continuously changing environment. They found that Meta-Learning is capable of adapting far more efficiently than baseline models in the few-shot regime. Although the implementation is outside the domain of Natural Language Processing, it is worth-noting that experts from different domains have implemented this method and sheds lights on authors to implement in NLU tasks. 

To sum up,  MAML and continual learning have been applied on NLP tasks separately but not both. In reinforcement learning, Meta-Continual learning can solve non-stationary environments \cite{Al-Shedivat:2017}. In next section, we extend on the work done by Javed and White \shortcite{MLRCL:19} and propose implementations on combining both methods for NLP tasks.

\section{Problem and Method}
\subsection{Problem Formation}
Consider the input data consists of an stream of data 
\begin{align*}
    \mathcal{T} = (X_1, Y_1), (X_2, Y_2), ...,(X_t, Y_t), ...    
\end{align*}for inputs $X_t$ and targets $Y_t$, for continual learning task this stream can be extended to an unending stream. In our case, we concatenate batches of data in order, each batch consisting data from a task in glue or superglue benchmark \shortcite{glue:19} \shortcite{superglue:19}. We followed Dou et al\shortcite{dou:19}'s way of defining Meta-training and Meta-testing tasks. For high resources tasks in Meta-Training, we use SST-2, QQP,1 MNLI and QNLI. For low-resource auxiliary tasks in Meta-testing, we add RTE, BoolQ, CB, Copa, WiC and WsC from SuperGlue to the original set of meta-testing tasks, CoLA , MRPC , STS-B and RTE.
\subsection{Method}
Javed and White at\shortcite{MLRCL:19} proposed a methodology that achieves Meta-Learning under continual learning setting. The representation learnt from existing knowledge by Meta-Learning, enables the model to learn new tasks quickly. Traditional MAML, proposed by Finn et al \shortcite{finn:18},takes a task $\mathcal{T}_{i}$ which is sampled from $p{(\mathcal{T})}$ during meta-training phrase and the model is trained with $K$ samples and feedback from the corresponding loss $L_{\mathcal{T}_i}$, and then tested on new samples selected from $\mathcal{T}_i$. Model is improved by looking at how test error on new data changes with respect to parameters. Finn's approach of MAML is learning an effective initialization which Javed et at\shortcite{MLRCL:19} reframed to MAML-Rep which enables it to work in the online setting. OML is another approach that attempts to alleviates catastrophic forgetting by online updating at meta-training phase, and utilize meta-testing tasks to improve the accuracy of the model in general. Given most neural networks are highly sparse, OML takes such advantage to update its parameters by constructing representations of the incoming online data point of different tasks either as parallel, where some updates can be beneficial for many tasks, or orthogonal, where updates by later task do not interfere with previous tasks. 


Our model architecture strictly follows the architecture proposed in \cite{MLRCL:19}, where both MAML-Rep and OML objectives are test in NLU tasks by training a pre-trained BERT model, we call models produced by these objectives MAML-Bert and OML-Bert. We used BERT because BERT is a state-of-art language model that utilizes Transformer architectures \cite{bert:18}. Pre-trained BERT is chosen instead of an empty BERT because Yogatama et al \shortcite{Yogatama:2019} have showed that training a BERT with supervised tasks instead of unsupervised tasks critically degrades model performance, and this paper focuses on supervised tasks only. To understand our training and evaluation methods, a brief overview of both objectives are introduced below.

For Meta-Training, we consider two Meta-Objective to minimize. (1) a MAML like objective and (2) OML objective. The OML objective is defined as 
\begin{align*}
    &\sum_{\mathcal{T}_i \sim p(\mathcal{T})} \text{OML}(W, \theta) \\=&  \sum_{\mathcal{T}_i \sim p(\mathcal{T})} \sum_{S_j^k \sim  p(S_k|\mathcal{T}_i)} \mathcal{L}(W, \theta, S_j^k)
\end{align*}
where $S_k^j = (X^i_{j+1}, Y^i_{j+1}), (X^i_{j+2}, Y^i_{j+2}),$...\\$,(X^i_{j+k}, Y^i_{j+k})$ sampled from distribution $p(S_k|\mathcal{T}_i)$. Specifically, we designed our \emph{mBERT} (modified BERT model ) inspired from \shortcite{MLRCL:19} into two parts: \emph{BERT-Base Network} (Representation Learning Network, RLN) and \emph{Task-specific Network} (Prediction Learning Network, PLN) illustrated in the Figure 1.\newline
\begin{figure}
    \centering
\includegraphics[width=\linewidth]{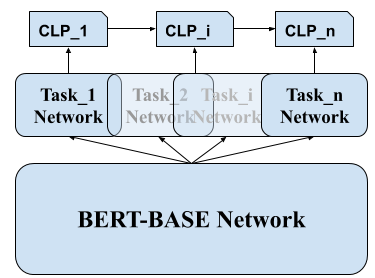}    \caption{mBERT (modified BERT) model for continual learning problems (CLPs)}
    \label{fig:my_label}
\end{figure}

Both approach updates PLN only in Meta-Training inner loop using support dataset with around 100 data points for some inner steps, and updates RLN in Meta-Trianing using query dataset with 80 data points in the outer loop. For Meta-Testing, 100 data points are fed into the model and both RLN and PLN is updated for some inner steps, at the end of each Meta-Tesing, we compare the performance of model trained on a specific task with after the model has been updated on all Meta-Testing Tasks to test if model forgets.

To minimize the objective above, our MAML-BERT model for continual learning are elaborated in Algorithm 1. For OML-BERT continual learning model, we proposed the same strategy except $step_8$ in \emph{Algorithm 1}, which was changed to train the inner update step with only one sample of the training data at a time. Using this method, we tried to mimic the Stochastic Gradient Descent approach in order to avoid catastrophic forgetting.
\begin{algorithm}[ht]
\SetAlgoLined
    \SetKwInOut{Input}{input}
    \SetKwInOut{Setting}{setting}
    \Input{CLP Tasks under distribution $p(\mathcal{T})$}
    \Setting{PLN (Classifier), RLN (BERT-Base) learning rates set as $\alpha, \beta$}
\nl Initialize \emph{BERT-Base} parameters $\theta$\;
 \nl \For{$T_i$ in $p(CLP)$}{
 \nl Fetch $T_i$ 's train dataset \;
 \nl Randomly Initialize \emph{$Task_i$ Classifier} parameters $W$\;
 \nl \For{j in number of inner update step}{
    \nl Freeze \emph{BERT-Base Network} \;
    \nl Train for $Batch_j$ of support data to update $W_j$
 }
 \nl Unfreeze \emph{BERT-Base Network} \;
 \nl Fetch $T_i$ 's test dataset \;
 \nl Update $\theta_i$ with updated $W_j$ \;
 }
\caption{MAML-BERT Training}

\end{algorithm}

\begin{algorithm}[ht]
\SetAlgoLined
    \SetKwInOut{Input}{input}
    \SetKwInOut{Setting}{setting}
    \Input{Target CLP Tasks under distribution $p(\mathcal{T})$}
    \Setting{Fine tuned learning rate $\alpha$ with Cosine Annealing Scheduler}
     \nl \For{$T_i$ in $p(Target Tasks)$}{
      \nl Randomly Initialize \emph{$Task_i$ Classifier} parameters $W$\;
      \nl \emph{BERT-Base} parameters $\theta$ from every 5 epoch Meta-training \;
       \nl \For{j in number of inner update step}{
       \nl fine-tuning both \emph{BERT-Base} and \emph{Classifier} \;
     \nl Metrics Calculation based on $T_i$ mode.
            }
\nl Test forgetting for previous Target CLP tasks $T_{0:i-1}$
}
    \caption{MAML-BERT Testing}
\end{algorithm}

\subsection{Evaluation}
We present our results along with Dou's and our baseline, using one model to simply train auxiliary tasks in sequences. With forgetting that degrades model performance, our MAML-Rep still outperforms other approaches. We attempted with different inner learning rates, inner update steps and Meta-Testing batch sizes, OML did not seem to generalize well in Meta-Testing tasks. Cola is evaluated with Matthew correlation, sts-b is evaluated with Pearson correlation, and rest two present accuracy score. 

\begin{center}
\small
  \resizebox{\columnwidth}{!}{\begin{tabular}{||c c c c c c ||} 
 \hline
Tasks & Dou et al & BERT & MAML-Rep & OML\\ [0.5ex] 
 \hline\hline
 CoLA &  \textbf{53.4} & 46.23/0 & 60.51/51.23 & 0/0 \\ 
 \hline
 MRPC  & 85.8& 75/56.25 &  \textbf{86.27/88.00} & 68.3/68.3\\
 \hline
 STS-B  &  \textbf{87.3} & 65.28/68.56 & 82.26/80.51 & 0/4.11\\
 \hline
RTE & 76.4& 56.25 &  \textbf{90.25} & 52.7\\
[1ex]
 \hline
\end{tabular}}
\end{center}

In addition, we swapped Meta-Testing tasks to low resource SuperGlue tasks and expanded number of tasks to 5 showing in table 2. All tasks are evaluated with accuracy score. MAML-Rep outperforms on three out of four tasks, note that OML still struggles to get better than random guessing for WsC tasks, and behave like random guessing for WiC and Copa tasks. 

\begin{center}
\small
 \begin{tabular}{|| c c c ||} 
 \hline
Tasks &  MAML-Rep & OML\\ [0.5ex] 
 \hline\hline
 WsC &  \textbf{75/72.11} & 36.53/36.53 \\ 
 \hline
 WiC  &  \textbf{52.98/53.29}&  50/50\\
 \hline
 BoolQ & 64.83/59.70& \textbf{61.82/62.17}\\
 \hline
 Copa &  \textbf{55/55}& 54.34/55\\
 \hline
 Cb & \textbf{ 100}&  89.12\\
 \hline
\end{tabular}
\end{center}


\section{Conclusion}
In this work, we are able to extend Meta-Learning under continual learning framework to learn a general presentation that is robust on a set of continual tasks with efficiency. We replicate \shortcite{MLRCL:19} method and and implement on NLU tasks. Results show that with less datapoints, we could derive a MAML like model that is robust on testing tasks, however extending it to continual setting during training phrase, the performance drastically worsen. Future direction would be extending this approach to other language models, as wells as experiment with a combination of high and low resources other than Glue and SuperGlue benchmark to evaluate model performance.



\bibliography{naaclhlt2019}
\bibliographystyle{acl_natbib}

\appendix

 \section{Appendices: Implementation Details}
\label{sec:appendix}
Our implementation is based on PyTorch implementation, backboned in Huggingface \textbf{$BERT_{BASE}$} model. We use Adam optimizer, with a batch size of 16 for both Meta-Training and Meta-Testing. Maximum sentence length is set to be 64. 
In Meta-Learning and Meta-Testiing stage, we use learning rate of $5e^{-5}$ for outer loop learning rate where we update RLN, and $5e^{-3}$ for inner learning to update PLN. We use a cosine annealing in Meta-Training as a scheduler to update the optimizer. Dropout of 0.1 is applied to PLN when it is applicable. We set the inner update step to 5 for Meta-Training and 7 for Meta-Testing. We use a total sample of 128 and 112 for support and query dataset for Meta-Testing, and 100 and entire dev set during Meta-Testing. 


\end{document}